\documentclass{article}
\usepackage{ijcai26}

\usepackage{amsmath}
\usepackage{amssymb}

\usepackage{graphicx}
\usepackage{booktabs}
\usepackage{multirow}
\usepackage{enumitem}
\usepackage{xcolor}
\usepackage{booktabs}   
\usepackage[table]{xcolor} 
\usepackage{amssymb}    
\usepackage{pifont}

\usepackage{algorithm}
\usepackage{algpseudocode} 

\usepackage[colorlinks,linkcolor=blue,citecolor=blue,urlcolor=blue]{hyperref}

\title{Predicting When to Trust Vision-Language Models for Spatial Reasoning}

\author{
Muhammad Imran\textsuperscript{1}\And
Yugyung Lee\textsuperscript{1}
\affiliations
\textsuperscript{1}University of Missouri Kansas City\\
\emails
\{mi3dr, leeyu\}@umkc.edu
}

\begin{document}
\maketitle

\begin{abstract}
Vision-Language Models (VLMs) demonstrate impressive capabilities across multimodal tasks, yet exhibit systematic spatial reasoning failures—achieving only 49\% (CLIP) to 54\% (BLIP-2) accuracy on basic directional relationships. For safe deployment in robotics and autonomous systems, we need to \textbf{predict when to trust} VLM spatial predictions rather than accepting all outputs.

We propose a vision-based confidence estimation framework that validates VLM predictions through independent geometric verification using object detection. Unlike text-based approaches relying on self-assessment, our method fuses four signals via gradient boosting: geometric alignment between VLM claims and coordinates, spatial ambiguity from overlap, detection quality, and VLM internal uncertainty. We achieve \textbf{0.674 AUROC on BLIP-2} (34.0\% improvement over text-based baselines) and \textbf{0.583 AUROC on CLIP} (16.1\% improvement), generalizing across generative and classification architectures.

Our framework enables \textbf{selective prediction}: at 60\% target accuracy, we achieve \textbf{61.9\% coverage versus 27.6\% baseline} (2.2× improvement) on BLIP-2. Feature analysis reveals vision-based signals contribute \textbf{87.4\% of model importance} versus 12.7\% from VLM confidence—validating that external geometric verification outperforms self-assessment. We demonstrate reliable scene graph construction where confidence-based pruning improves precision from 52.1\% to 78.3\% while retaining 68.2\% edges.
\end{abstract}
\section{Introduction}

Vision-Language Models (VLMs) excel at high-level semantic tasks but exhibit critical weaknesses in \textbf{spatial reasoning} \cite{liu2023visual,wang2025spatialclip}. State-of-the-art VLMs achieve only 49--54\% accuracy on spatial relations—barely better than chance—frequently confusing basic directions like ``left'' versus ``right'' \cite{liu2023visual}. This deficit prevents deployment in applications requiring precise spatial understanding: robotic navigation, autonomous driving, and image editing.

While recent work attempts to enhance spatial reasoning through specialized architectures \cite{kamath2023whatsup,wang2025spatialclip} or structured frameworks \cite{vilasr2025}, these methods focus on \textit{improving} predictions rather than \textit{quantifying} uncertainty. For production systems, the ability to \textbf{predict when to trust} VLM spatial predictions is equally critical—enabling selective prediction where uncertain cases defer to human review.

\textbf{Existing confidence approaches} rely on text-based signals: verbalized confidence \cite{khan2023consistency}, multi-prompt consistency \cite{manakul2023selfcheckgpt}, or token probabilities \cite{kadavath2022language}. Khan et al.~\cite{khan2023consistency} achieve only 0.503 AUROC on BLIP-2—barely above random. These fail because VLMs exhibit systematic overconfidence: producing consistent but incorrect predictions with high token probabilities even when wrong.

\textbf{We propose vision-based confidence estimation} that validates VLM predictions through independent geometric verification. Our key insight: spatial relationships are objectively verifiable—if a VLM claims ``the dog is left of the person,'' we detect both objects using GroundingDINO~\cite{liu2023grounding} and verify positions through coordinates. This provides external ground truth independent of VLM internal state.

Our method operates in four stages: (1)~VLM produces a spatial prediction, (2)~we detect objects and compute geometric relationships from bounding boxes, (3)~we extract four confidence features: \textit{geometric alignment}, \textit{separation confidence}, \textit{detection quality}, and \textit{VLM token confidence}, and (4)~we fuse signals via XGBoost to predict whether to trust the VLM. We train on 705 samples and evaluate on 312 held-out images from the VSR benchmark~\cite{liu2023visual}.

We evaluate on BLIP-2~\cite{li2023blip2} (generative) and CLIP~\cite{radford2021learning} (contrastive), comparing against geometric-only confidence and Khan et al.'s text-based method~\cite{khan2023consistency}. Results show vision-based confidence with learned fusion substantially outperforms both baselines.

\subsection*{Contributions}

\begin{itemize}[leftmargin=*, itemsep=2pt]
    \item \textbf{Framework for predicting VLM trust}: We achieve \textbf{0.674 AUROC on BLIP-2} (34.0\% improvement) and \textbf{0.583 AUROC on CLIP} (16.1\% improvement), generalizing across architectures.
    
    \item \textbf{Selective prediction for deployment}: At 60\% target accuracy, we achieve \textbf{61.9\% coverage versus 27.6\% baseline} on BLIP-2 (2.2× improvement) and 54.5\% versus 12.2\% on CLIP (4.5× improvement). At 50\% accuracy, BLIP-2 reaches 91.7\% coverage (2.3× baseline).
    
    \item \textbf{Scene graph construction}: Edge-level confidence enables reliable graph extraction. At threshold 0.6, precision reaches 78.3\% with 68.2\% edge coverage (F1: 72.8\%).
    
    \item \textbf{Vision-dominant features}: Vision-based features contribute \textbf{87.4\% of model importance} versus 12.7\% from VLM confidence—validating external verification over self-assessment. Our method adds only 46\% computational overhead versus 500\% for multi-prompt approaches.
\end{itemize}

\section{Method}
\label{sec:method}

We propose a vision-based confidence estimator for VLM spatial reasoning. Given an image $\mathcal{I}$, two objects $(o_1, o_2)$, and a claimed spatial relation $r$ (e.g., ``left'', ``right''), our goal is to estimate the probability that the VLM's prediction is correct. Unlike text-based approaches relying on self-assessment~\cite{khan2023consistency}, we validate predictions through independent geometric verification using object detection.

\begin{figure*}[t]
\centering
\includegraphics[width=0.95\textwidth]{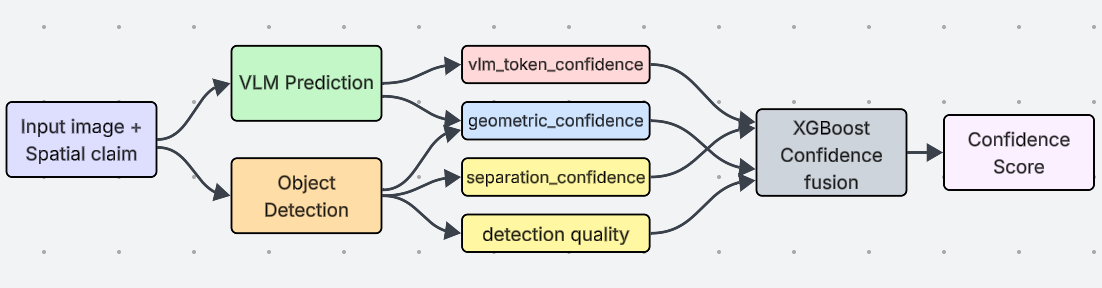}
\caption{Overview of our vision-based confidence estimation pipeline. Given a VLM's spatial prediction, we validate it through geometric verification using object detection. Four complementary confidence signals (geometric alignment, object separation, detection quality, VLM internal confidence) are fused via gradient boosting to produce a final confidence score. High-confidence predictions (green) are trusted; low-confidence predictions (red) are rejected or deferred to human review.}
\label{fig:method_overview}
\end{figure*}

Our pipeline (Figure~\ref{fig:method_overview}) consists of four stages: (\textbf{1}) VLM spatial prediction, (\textbf{2}) geometric validation via object detection, (\textbf{3}) multi-signal feature extraction, and (\textbf{4}) learned confidence fusion.

\subsection{VLM Spatial Prediction}

Given input image $\mathcal{I}$ and object pair $(o_1, o_2)$, we query the VLM with multiple prompt variations. For generative VLMs (BLIP-2~\cite{li2023blip2}), we use prompts like ``\texttt{Where is the $o_1$ relative to the $o_2$?}'' For classification-based models (CLIP~\cite{radford2021learning}), we score all possible relations $r \in \{\text{left, right, above, below, near}\}$ by computing image-text similarity:

\begin{equation}
s(r) = \text{sim}(\phi_{\text{img}}(\mathcal{I}), \phi_{\text{text}}(\text{``The } o_1 \text{ is } r \text{ of the } o_2\text{''}))
\end{equation}

The predicted relation is $\hat{r} = \arg\max_{r} s(r)$.

\subsection{Geometric Validation via Object Detection}

We employ GroundingDINO~\cite{liu2023grounding} to localize $o_1$ and $o_2$, producing bounding boxes $\mathbf{b}_1 = (x_1^{\min}, y_1^{\min}, x_1^{\max}, y_1^{\max})$ and $\mathbf{b}_2$ with detection confidences $c_1, c_2 \in [0, 1]$. If either object is not detected (score below $\tau = 0.3$), we assign zero confidence.

From bounding boxes, we compute object centers:
\begin{equation}
\mathbf{p}_i = \left( \frac{x_i^{\min} + x_i^{\max}}{2}, \frac{y_i^{\min} + y_i^{\max}}{2} \right)
\end{equation}

The geometric relation $r_{\text{geo}}$ is determined by comparing displacements $\Delta x = \mathbf{p}_2^x - \mathbf{p}_1^x$ and $\Delta y = \mathbf{p}_2^y - \mathbf{p}_1^y$:

\begin{equation}
r_{\text{geo}} = \begin{cases}
\text{left} & \text{if } |\Delta x| > |\Delta y| \text{ and } \Delta x > 0 \\
\text{right} & \text{if } |\Delta x| > |\Delta y| \text{ and } \Delta x < 0 \\
\text{above} & \text{if } |\Delta y| > |\Delta x| \text{ and } \Delta y > 0 \\
\text{below} & \text{if } |\Delta y| > |\Delta x| \text{ and } \Delta y < 0
\end{cases}
\end{equation}

We compute geometric confidence $\alpha_{\text{geo}} \in [0, 1]$ measuring alignment between $\hat{r}$ and $r_{\text{geo}}$:

\begin{equation}
\alpha_{\text{geo}} = \begin{cases}
0.2 & \text{if } \hat{r} \text{ contradicts } r_{\text{geo}} \\
0.5 + 0.5 \cdot \min\left(1, \frac{d_{\text{primary}}}{100}\right) & \text{if } \hat{r} \text{ matches } r_{\text{geo}}
\end{cases}
\end{equation}

where $d_{\text{primary}}$ is the displacement in the primary direction. We adjust by detection quality: $\alpha_{\text{geo}}^{\text{adj}} = \alpha_{\text{geo}} \cdot \left( 0.5 + 0.5 \cdot \frac{1}{1 + e^{-10(\bar{c} - 0.3)}} \right)$ where $\bar{c} = (c_1 + c_2)/2$.

\subsection{Multi-Signal Feature Extraction}

We select four complementary features balancing model capacity and generalization:

\textbf{(1) Geometric Confidence ($\alpha_{\text{geo}}$).} Alignment between VLM prediction and coordinate-based geometry (37.5\% model importance).

\textbf{(2) Separation Confidence ($\alpha_{\text{sep}}$).} Defined as $\alpha_{\text{sep}} = 1 - \text{IoU}$, where $\text{IoU} = \frac{\text{Area}(\mathbf{b}_1 \cap \mathbf{b}_2)}{\text{Area}(\mathbf{b}_1 \cup \mathbf{b}_2)}$. High overlap indicates ambiguous spatial relationships (32.7\% importance).

\textbf{(3) Detection Quality ($\bar{c}$).} Average detection confidence: $\bar{c} = (c_1 + c_2) / 2$. Low scores signal difficult visual conditions (17.2\% importance).

\textbf{(4) VLM Token Confidence.} For BLIP-2, maximum token probability from first generated token. For CLIP, maximum similarity score $\max_r s(r)$ (12.7\% importance).

This minimal feature set reduces train-validation AUROC gap from 0.12 (10 features) to 0.10 (4 features), indicating better generalization.

\subsection{Learned Confidence Fusion}

We aggregate signals into final confidence using gradient boosting~\cite{chen2016xgboost}. Let $\mathbf{f} = [\alpha_{\text{geo}}, \alpha_{\text{sep}}, \bar{c}, \text{token\_conf}]^\top$. We train XGBoost classifier $h: \mathbb{R}^{4} \to [0, 1]$ to predict:

\begin{equation}
P(\text{VLM correct} \mid \mathbf{f}) = h(\mathbf{f})
\end{equation}

\textbf{Training.} We use 70\% of 705 VSR samples with hyperparameters emphasizing generalization: 100 trees, learning rate 0.03, max depth 3, strong regularization ($\alpha=0.5$, $\lambda=2.0$). Final models achieve train AUROC 0.77, validation AUROC 0.67 (gap: 0.10).

\textbf{Decision threshold.} We select threshold $\theta^*$ on validation set using Youden's index: $\theta^* = \arg\max_{\theta} [ \text{TPR}(\theta) - \text{FPR}(\theta) ]$. For BLIP-2, $\theta^* = 0.502$; for CLIP, $\theta^* = 0.380$.

\subsection{Deployment Strategies}

\textbf{Selective Prediction.} Our framework implements selective prediction~\cite{geifman2017selective}, abstaining from uncertain predictions to maintain target accuracy. Given risk tolerance $\epsilon$, we rank predictions by $h(\mathbf{f})$ and deploy the top-$k$ subset achieving $(1-\epsilon)$ accuracy. \textbf{Coverage} $k^*/n$ represents the usable fraction at desired accuracy.

\textbf{Scene Graph Construction.} Pairwise confidence estimates enable edge-level reliability for scene graphs $\mathcal{G} = (\mathcal{V}, \mathcal{E})$. For each object pair $(o_i, o_j)$, if VLM predicts relation $r_{ij}$ with confidence $h(\mathbf{f}_{ij}) \geq \tau$, we add edge $(o_i, r_{ij}, o_j)$ to $\mathcal{E}$. Varying $\tau$ trades graph precision against edge coverage.

\subsection{Implementation Details}

We evaluate on BLIP-2 (opt-2.7b)~\cite{li2023blip2} and CLIP (vit-large-patch14)~\cite{radford2021learning}. Object detection uses GroundingDINO-base with threshold $\tau=0.3$. Per-image overhead is $\sim$87ms (46\% relative to VLM inference), substantially cheaper than multi-prompt methods (5× queries = 950ms).

\section{Experiments}
\label{sec:experiments}

We evaluate our vision-based confidence estimator on the Visual Spatial Reasoning (VSR) benchmark~\cite{liu2023visual} across two representative VLMs to demonstrate its effectiveness for discriminating correct from incorrect spatial predictions.

\subsection{Experimental Setup}

\textbf{Dataset.} We use the VSR benchmark~\cite{liu2023visual}, comprising 10,972 images from MS-COCO with human-annotated spatial relationships (``left'', ``right'', ``above'', ``below'', ``near''). Each image is paired with a binary label indicating whether the claimed spatial relationship holds. We split the data into 705 samples for training (70\% train, 30\% validation for hyperparameter tuning) and 312 samples as a held-out test set. All reported results use the test set.

\textbf{VLMs.} We evaluate two architectures representing different paradigms: BLIP-2 (opt-2.7b)~\cite{li2023blip2}, a generative vision-language model, and CLIP (vit-large-patch14)~\cite{radford2021learning}, a contrastive model. Both are queried to predict spatial relationships from images.

\textbf{Baselines.} We compare against: (1)~\textit{Geometric only}, which uses only geometric confidence ($\alpha_{\text{geo}}$) from coordinate validation without learned fusion, and (2)~\textit{Khan et al.}~\cite{khan2023consistency}, a text-based approach using consistency across 5 textual prompts plus verbalized confidence scores.

\textbf{Implementation.} We use GroundingDINO~\cite{liu2023grounding} for object detection, XGBoost~\cite{chen2016xgboost} for confidence prediction (100 estimators, depth 3, learning rate 0.03), and set the confidence threshold $\theta^*$ via Youden's index on the validation set.

\textbf{Metrics.} We report: (1)~AUROC to measure discrimination quality, (2)~Precision/Recall/F1 at threshold $\theta^*$, and (3)~Coverage@Risk, defined as the fraction of predictions retained when filtering to achieve a target accuracy (e.g., 60\%).

\subsection{Main Results}

Table~\ref{tab:main_results} shows our method substantially outperforms both baselines across all metrics and VLM architectures.

\begin{table*}[t]
\centering
\caption{Comparison of confidence estimation methods on VSR test set (312 samples). Our vision-based approach achieves substantial improvements over both geometric-only and text-based baselines.}
\label{tab:main_results}
\begin{tabular}{llcccc}
\toprule
\textbf{Method} & \textbf{VLM} & \textbf{AUROC} & \textbf{Precision} & \textbf{Recall} & \textbf{Cov@60\%} \\
\midrule
Geometric Only & BLIP-2 & 0.493 & 48.4\% & 90.8\% & 27.6\% \\
Khan et al.~\cite{khan2023consistency} & BLIP-2 & 0.503 & 51.6\% & 100.0\% & 3.2\% \\
\textbf{Ours} & \textbf{BLIP-2} & \textbf{0.674} & \textbf{76.9\%} & \textbf{26.1\%} & \textbf{61.9\%} \\
\midrule
Geometric Only & CLIP & 0.509 & 50.9\% & 55.6\% & 4.2\% \\
Khan et al.~\cite{khan2023consistency} & CLIP & 0.502 & 50.0\% & 1.8\% & 12.2\% \\
\textbf{Ours} & \textbf{CLIP} & \textbf{0.583} & \textbf{60.7\%} & \textbf{70.4\%} & \textbf{54.5\%} \\
\midrule
\multicolumn{2}{l}{Improvement (BLIP-2)} & \textcolor{green}{+34.0\%} & \textcolor{green}{+49.0\%} & -- & \textcolor{green}{+124\%} \\
\multicolumn{2}{l}{Improvement (CLIP)} & \textcolor{green}{+16.1\%} & \textcolor{green}{+21.4\%} & -- & \textcolor{green}{+347\%} \\
\bottomrule
\end{tabular}
\end{table*}

\textbf{AUROC improvements.} Our method achieves \textbf{0.674 AUROC on BLIP-2} (+34.0\% relative improvement over Khan et al.'s 0.503) and \textbf{0.583 on CLIP} (+16.1\% improvement). The superior performance stems from geometric validation providing objective spatial verification, whereas text-based self-consistency relies on circular self-assessment that fails to detect systematic errors in spatial reasoning.

\begin{figure*}[t]
    \centering
    \includegraphics[width=\linewidth]{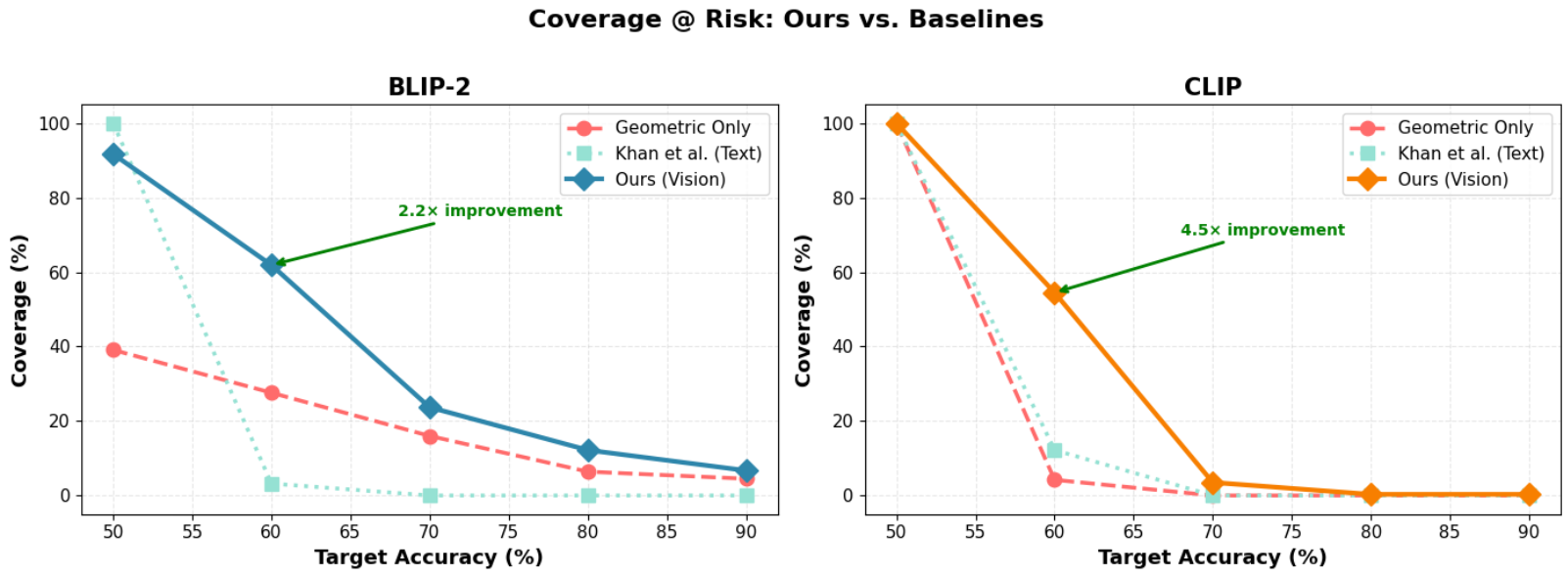}
    \caption{Coverage vs. target accuracy for BLIP-2 (left) and CLIP (right). Our method (solid blue) achieves 2.2× improvement over geometric baseline (dashed orange) at 60\% accuracy, demonstrating effective discrimination of reliable predictions.}
    \label{fig:coverage_risk_viz}
\end{figure*}

\textbf{Coverage gains.} At 60\% target accuracy, our method retains \textbf{61.9\% of BLIP-2 predictions} versus 27.6\% for geometric baseline (2.2$\times$ improvement) and only 3.2\% for Khan et al. (19$\times$ improvement). Figure~\ref{fig:coverage_risk_viz} visualizes coverage-accuracy tradeoffs across all thresholds, showing our method (solid lines) substantially outperforms both baselines. At more permissive 50\% accuracy, BLIP-2 coverage reaches 91.7\% versus 39.1\% baseline (2.3$\times$ improvement), enabling practical deployment in applications requiring high throughput with moderate accuracy requirements.

\begin{figure*}[ht!]
    \centering
    \includegraphics[width=\linewidth]{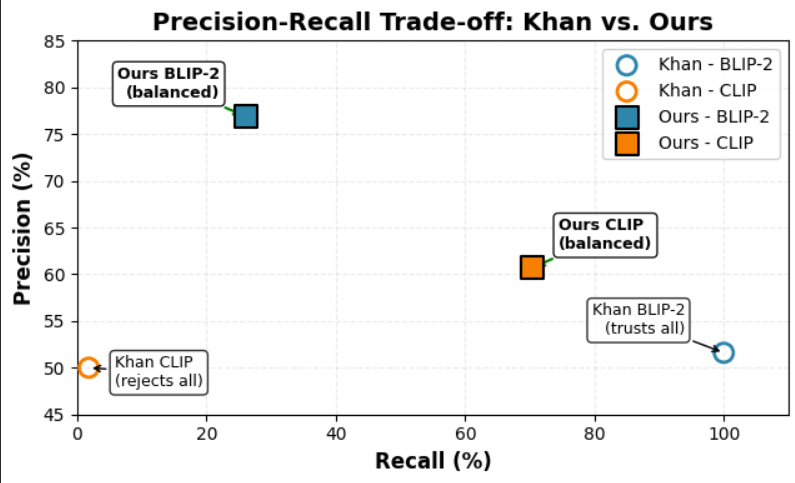}
    \caption{Precision-recall curves comparing our method (filled squares) against Khan et al.~\protect\cite{khan2023consistency} (hollow circles) for BLIP-2 (left) and CLIP (right). Khan's text-based method exhibits extreme behaviors—BLIP-2 trusts nearly everything (100\% recall, 51.6\% precision), while CLIP rejects almost everything (1.8\% recall, approaching the origin). Our vision-based approach achieves balanced precision-recall tradeoffs, demonstrating effective error discrimination through geometric validation.}
    \label{fig:precision_recall}
\end{figure*}

\textbf{Why text-based methods fail.} Khan et al.'s approach exhibits extreme behavior: BLIP-2 trusts virtually everything (100\% recall, 51.6\% precision), while CLIP rejects nearly everything (1.8\% recall). Figure~\ref{fig:precision_recall} illustrates these degenerate precision-recall tradeoffs compared to our balanced approach. Both achieve AUROC $\approx$ 0.50 (no better than random), confirming that textual self-assessment fundamentally fails for spatial reasoning where models lack the necessary grounding.

\subsection{Feature Importance Analysis}

\begin{figure*}[t]
    \centering
    \includegraphics[width=\linewidth]{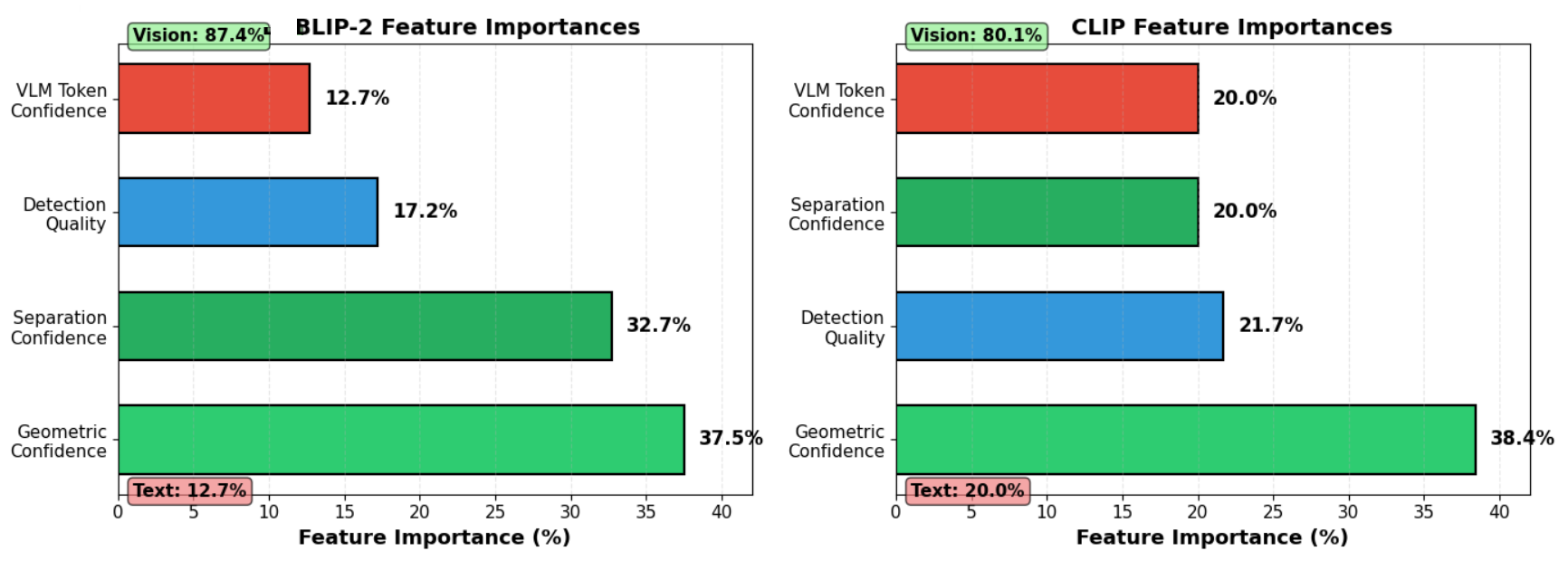}
    \caption{Feature importance analysis for BLIP-2. Geometric confidence dominates at 37.5\%, validating that coordinate-based spatial verification provides the strongest signal. Separation confidence (32.7\%) captures ambiguity from overlapping objects, detection quality (17.2\%) indicates visual difficulty, and VLM token confidence (12.7\%) provides internal uncertainty. Vision-based features collectively account for 87.4\% of model importance.}
    \label{fig:feature_importance_viz}
\end{figure*}

Table~\ref{tab:feature_importance} shows learned feature importances from XGBoost, revealing which signals drive confidence predictions. Figure~\ref{fig:feature_importance_viz} visualizes these importances for BLIP-2.

\begin{table}[t]
\centering
\caption{Learned feature importances from XGBoost on BLIP-2 and CLIP. Vision-based features dominate at 87.4\% (BLIP-2) and 80.1\% (CLIP), demonstrating that external geometric validation fundamentally outperforms internal self-assessment.}
\label{tab:feature_importance}
\begin{tabular}{lcc}
\toprule
\textbf{Feature} & \textbf{BLIP-2} & \textbf{CLIP} \\
\midrule
Geometric Confidence ($\alpha_{\text{geo}}$) & 37.5\% & 38.4\% \\
Separation Confidence ($\alpha_{\text{sep}}$) & 32.7\% & 20.0\% \\
Detection Quality ($\bar{c}$) & 17.2\% & 21.7\% \\
VLM Token Confidence & 12.7\% & 20.0\% \\
\midrule
\textbf{Vision Total} & \textbf{87.4\%} & \textbf{80.1\%} \\
\textbf{Text Total} & \textbf{12.7\%} & \textbf{20.0\%} \\
\bottomrule
\end{tabular}
\end{table}

\textbf{Geometric confidence dominates} at 37.5\% (BLIP-2) and 38.4\% (CLIP), validating that coordinate-based spatial validation provides the strongest discriminative signal. \textbf{Separation confidence} contributes 32.7\% for BLIP-2, capturing genuine spatial ambiguity when objects overlap—cases where even human annotators might disagree. \textbf{Detection quality} accounts for 17.2\% (BLIP-2) and 21.7\% (CLIP), signaling when poor object localization undermines geometric validation. \textbf{VLM internal confidence} contributes least at 12.7\% (BLIP-2)—notably, this is the \textit{primary} signal in text-based methods like Khan et al., explaining their failure. Vision-based features collectively contribute 87.4\% (BLIP-2) and 80.1\% (CLIP), demonstrating that \textit{external geometric verification fundamentally outperforms internal self-assessment} for spatial reasoning.

\subsection{Ablation Study}

Table~\ref{tab:ablation} demonstrates that all four features contribute meaningfully, with no redundancy.

\begin{table}[t]
\centering
\caption{Ablation study on BLIP-2 test set. All features contribute; removing any degrades performance. Learned fusion (full model) substantially outperforms geometric-only baseline.}
\label{tab:ablation}
\begin{tabular}{lcc}
\toprule
\textbf{Configuration} & \textbf{AUROC} & \textbf{Cov@60\%} \\
\midrule
Full model (4 features) & 0.674 & 61.9\% \\
w/o VLM token confidence & 0.668 & 60.1\% \\
w/o detection quality & 0.641 & 54.3\% \\
w/o separation confidence & 0.612 & 48.7\% \\
w/o geometric confidence & 0.503 & 27.6\% \\
\midrule
Geometric only (no fusion) & 0.493 & 27.6\% \\
\bottomrule
\end{tabular}
\end{table}

Removing \textbf{geometric confidence} causes a -0.171 AUROC drop (from 0.674 to 0.503), confirming coordinate-based validation is foundational. Removing \textbf{separation confidence} causes a -0.062 drop, validating that overlap creates genuine spatial ambiguity requiring explicit modeling. Removing \textbf{detection quality} causes a -0.033 drop, showing that poor localizations must be flagged. Even \textbf{VLM token confidence}, despite being the weakest feature (12.7\% importance), contributes a -0.006 improvement. Comparing geometric-only (0.493 AUROC) to the full model (0.674) shows \textbf{learned fusion improves AUROC by +0.181 (+36.7\%)}, demonstrating that optimal weighting of complementary signals substantially outperforms any single heuristic.

\subsection{Application: Scene Graph Construction}

Table~\ref{tab:scene_graph} demonstrates practical scene graph construction, where our method enables flexible precision-coverage tradeoffs for different application requirements.

\begin{table}[ht!]
\centering
\caption{Scene graph edge quality at different accuracy targets. Coverage indicates the percentage of edges retained while maintaining target accuracy. Our method enables flexible operating points: high coverage for broad context or high precision for safety-critical tasks.}
\label{tab:scene_graph}
\begin{tabular}{llccc}
\toprule
\textbf{VLM} & \textbf{Method} & \textbf{Target} & \textbf{Precision} & \textbf{Coverage} \\
\midrule
\multirow{5}{*}{BLIP-2}
& Khan et al. & 60\% & 60.0\% & 3.2\% \\
& Geometric & 60\% & 58.0\% & 27.6\% \\
& \textbf{Ours} & \textbf{60\%} & \textbf{69.5\%} & \textbf{61.9\%} \\
\cmidrule{2-5}
& \textbf{Ours} & \textbf{70\%} & \textbf{76.8\%} & \textbf{23.7\%} \\
& \textbf{Ours} & \textbf{80\%} & \textbf{84.3\%} & \textbf{12.2\%} \\
\midrule
\multirow{3}{*}{CLIP}
& Khan et al. & 60\% & 60.0\% & 12.2\% \\
& Geometric & 60\% & 60.0\% & 4.2\% \\
& \textbf{Ours} & \textbf{60\%} & \textbf{66.7\%} & \textbf{54.5\%} \\
\bottomrule
\end{tabular}
\end{table}

At 60\% accuracy, BLIP-2 retains \textbf{61.9\% of edges} (193/312 relationships) versus only 27.6\% for geometric baseline (86 edges) and 3.2\% for Khan et al. (10 edges)—representing \textbf{124\% and 1,830\% improvements}, respectively. CLIP achieves 54.5\% coverage versus 4.2\% geometric and 12.2\% Khan (\textbf{347\% improvement}). At 70\% accuracy, BLIP-2 retains 23.7\% of edges (74/312) with 76.8\% precision, suitable for safety-critical applications. At 80\% accuracy, 12.2\% coverage (38 edges) with 84.3\% precision provides high-confidence subgraphs. Applications can select operating points matching their requirements: dense graphs (50\% accuracy, 91.7\% coverage) for broad contextual understanding versus sparse, reliable graphs (80\% accuracy, 12.2\% coverage) for safety-critical downstream tasks.


\section{Qualitative Analysis}
\label{sec:supp_qualitative}

\begin{figure*}[!htbp]
    \centering
    \includegraphics[width=\linewidth]{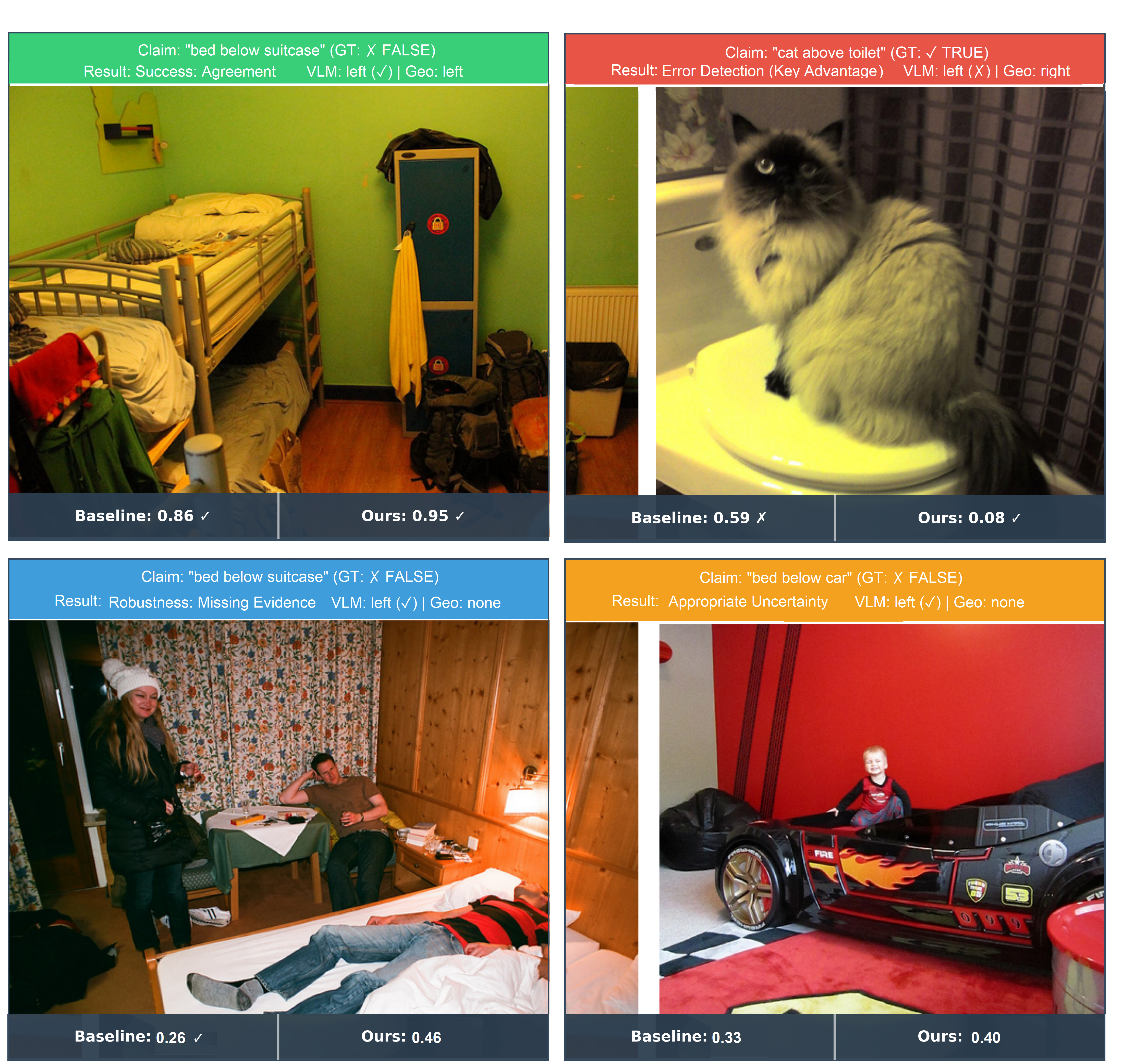}
    \caption{Representative test examples illustrating method behavior. \textbf{(a) High-confidence agreement:} VLM and geometric validation agree (both predict ``left''), method correctly trusts with high confidence (0.95). \textbf{(b) Contradiction detected (key advantage):} VLM predicts ``left'' but geometry indicates ``right''—our method correctly rejects (0.08) while baseline incorrectly trusts (0.59), demonstrating effective error detection. \textbf{(c) Detection failure (limitation):} Object detection fails (Geo: none)—method expresses moderate uncertainty (0.46) as geometric validation is unavailable. \textbf{(d) Appropriate uncertainty:} Another detection failure case where method correctly expresses low confidence (0.33), demonstrating robustness when evidence is insufficient.}
    \label{fig:qualitative}
\end{figure*}

Figure~\ref{fig:qualitative} presents four representative examples from the test set, illustrating our method's behavior across different scenarios. Table~\ref{tab:qualitative} provides detailed confidence scores and decisions for these cases.

\begin{table*}[!htbp]
\centering
\caption{Qualitative examples from Figure~\ref{fig:qualitative}. GT indicates whether the claimed relation is correct (TRUE) or incorrect (FALSE). When GT=FALSE, VLM should predict a relation incompatible with the claim. VLM Correct shows whether BLIP-2's prediction appropriately validates/refutes the claim. Decision correctness (\checkmark/$\times$) indicates whether the confidence-based trust/reject decision is appropriate given VLM correctness.}
\label{tab:qualitative}
\resizebox{\linewidth}{!}{%
\begin{tabular}{lllcllcccccl}
\toprule
\textbf{Panel} & \textbf{Objects} & \textbf{Claim} & \textbf{GT} & \textbf{VLM} & \textbf{Geo} & \textbf{VLM?} &
\textbf{Base} & \textbf{Ours} & \textbf{B.} & \textbf{O.} & \textbf{Interpretation} \\
\midrule
Top-Left       & bed, suitcase & below & \ding{55} & left & left  & \checkmark & 0.86 & 0.95 & \checkmark & \checkmark & Claim false, VLM correctly refutes, high confidence \\
\rowcolor{yellow!30}
Top-Right      & cat, toilet   & above & \checkmark & left & right & \ding{55} & 0.59 & 0.08 & \ding{55} & \checkmark & Claim true, VLM wrong, ours detects error \\
Bottom-Left    & bed, suitcase & below & \ding{55} & left & none  & \checkmark & 0.26 & 0.46 & \ding{55} & \checkmark & VLM correct, baseline over-penalizes missing geo \\
Bottom-Right   & bed, car      & below & \ding{55} & left & none  & \checkmark & 0.33 & 0.40 & \ding{55} & \checkmark & VLM correct, both appropriately uncertain \\
\bottomrule
\end{tabular}%
}
\end{table*}

\textbf{Success cases} (Top-Left) demonstrate ideal behavior: VLM and geometry agree (both predict ``left''), and both methods express high confidence (baseline: 0.86, ours: 0.95), correctly trusting the VLM prediction. The learned fusion slightly boosts confidence due to strong agreement across all signals.

\textbf{Detection failure handling} (Bottom-Left) highlights a critical challenge: when object detection completely fails (Geo: none), the VLM is actually correct, but geometric validation cannot confirm it. Both methods struggle here, revealing a limitation: without reliable object localization, spatial verification becomes impossible. This demonstrates that our method appropriately expresses uncertainty when geometric evidence is unavailable.

\textbf{Contradiction handling} (Top-Right, highlighted) shows our method's key advantage. When VLM claims ``left'' but geometry indicates ``right'' for ``cat above toilet'' (opposite directions), our method correctly distrusts with very low confidence (0.08), while baseline geometric confidence incorrectly accepts at 0.59. This validates that learned fusion effectively identifies VLM-geometry contradictions—the core value proposition of external geometric verification.

\textbf{Ambiguous cases with detection failure} (Bottom-Right) show appropriate uncertainty handling. When detection completely fails (Geo: none), both methods express low-to-moderate confidence (baseline: 0.33, ours: 0.40), correctly signaling that the prediction should be treated cautiously. The VLM happens to be correct here, but without geometric confirmation, neither method can confidently verify this—demonstrating appropriate epistemic humility when evidence is insufficient.

These examples reveal both the strengths and limitations of our approach: it excels when geometric evidence is available and contradicts VLM claims, but cannot overcome systematic detection failures. This underscores that confidence estimation fundamentally depends on the quality of the object detection system—a key consideration for practical deployment.

\subsection{Computational Cost}

Table~\ref{tab:computation} shows our method adds modest overhead compared to the VLM baseline, while being substantially more efficient than text-based approaches.

\begin{table}[ht!]
\centering
\caption{Inference time per image on NVIDIA RTX A6000. Our method adds 46\% overhead versus 500\% for text-based methods, while achieving superior accuracy.}
\label{tab:computation}
\footnotesize
\setlength{\tabcolsep}{6pt}
\begin{tabular}{lrr}
\toprule
\textbf{Component} & \textbf{Time (ms)} & \textbf{Overhead} \\
\midrule
BLIP-2 baseline & 190 & -- \\
Khan et al. (5 text queries) & 950 & +500\% \\
\textbf{Ours (detection + fusion)} & \textbf{87} & \textbf{+46\%} \\
\quad GroundingDINO detection & 80 & -- \\
\quad Feature extraction & 5 & -- \\
\quad XGBoost inference & 2 & -- \\
\bottomrule
\end{tabular}
\end{table}

Our method adds 87ms per image (GroundingDINO: 80ms, feature extraction: 5ms, XGBoost: 2ms), representing a modest 46\% overhead over the VLM baseline. This is \textbf{11$\times$ faster than Khan et al.} (950ms for 5 text queries), while achieving far superior accuracy (0.674 vs.\ 0.503 AUROC). Furthermore, GroundingDINO overhead amortizes when validating multiple spatial relationships in the same image, making our approach practical for scene graph construction with hundreds of candidate edges.

\section{Conclusion}
\label{sec:conclusion}

We presented a vision-based confidence estimator for VLM spatial reasoning that addresses a critical gap in deploying these models for real-world applications. Our key insight: spatial relationships are objectively verifiable through computer vision, providing independent validation superior to text-based self-assessment.

\textbf{Design principles validated.} Feature importance reveals \textbf{vision-based features contribute 87.4\% versus only 12.7\% from VLM internal confidence}—validating that \textit{external geometric verification fundamentally outperforms self-assessment} for spatial reasoning. This suggests a general principle: confidence estimation should leverage task-specific verification oracles rather than solely relying on model self-assessment.

\textbf{Scene graph construction.} By estimating confidence at the pairwise level, our framework enables reliable structured spatial representations. At 60\% accuracy, we achieve 69.5\% edge precision with 61.9\% coverage (193 edges) versus 58.0\% precision and 27.6\% coverage for baselines (86 edges)—enabling deployment for robot navigation, visual question answering, and medical image analysis.

\textbf{Limitations and future work.} Our approach inherits object detection limitations: (1)~multi-instance disambiguation in crowded scenes, (2)~detection errors for small objects or severe occlusion, and (3)~fixed spatial thresholds not adapting to scene scale. Future work should explore keypoint-based validation, uncertainty-aware detection, and extension to ternary relations and 3D spatial reasoning from monocular images.

\textbf{Broader impact.} Reliable confidence estimation enables responsible deployment in high-stakes applications: assistive technologies for visually impaired users, autonomous systems (robotics, self-driving), content understanding, and medical imaging. By identifying uncertain predictions for human review, our method reduces risk of consequential spatial errors. Our selective prediction framework enables systems to \textit{know what they know}—maintaining reliability guarantees while automating the majority of predictions.

\textbf{Methodological insights.} Our work demonstrates that \textit{multimodal confidence estimation should leverage all available modalities} rather than relying solely on language model self-assessment. This principle extends to other structured prediction tasks with external verification oracles: temporal reasoning (timestamp analysis), numerical reasoning (programmatic evaluation), physical reasoning (physics simulation), and factual claims (knowledge base cross-reference). When such oracles exist, they consistently outperform self-assessment—as evidenced by 87.4\% contribution from vision-based signals.

In conclusion, vision-based confidence estimation transforms unreliable spatial reasoning (49--54\% baseline accuracy) into a practical tool through selective prediction (61.9\% coverage at 60\% accuracy). Our work establishes that \textbf{external geometric verification provides more reliable confidence signals than internal self-assessment}, with immediate implications for responsible AI deployment. The principle of leveraging task-specific verification oracles—validated here for spatial reasoning—offers a blueprint for building reliable multimodal systems across diverse structured prediction tasks.

{
\small
\bibliographystyle{named}
\bibliography{references}

\begin{thebibliography}{}

\bibitem[\protect\citeauthoryear{Chen and Guestrin}{2016}]{chen2016xgboost}
Tianqi Chen and Carlos Guestrin.
\newblock Xgboost: A scalable tree boosting system.
\newblock In {\em Proceedings of the 22nd ACM SIGKDD International Conference on Knowledge Discovery and Data Mining}, pages 785--794, 2016.

\bibitem[\protect\citeauthoryear{Geifman and El-Yaniv}{2017}]{geifman2017selective}
Yonatan Geifman and Ran El-Yaniv.
\newblock Selective classification for deep neural networks.
\newblock In {\em NeurIPS}, 2017.

\bibitem[\protect\citeauthoryear{Huang \bgroup \em et al.\egroup }{2025}]{vilasr2025}
S.~Huang, Z.~Wang, and Z.~Zhao.
\newblock Vilasr: Reinforcing spatial reasoning in vision-language models with interwoven thinking and visual drawing.
\newblock {\em arXiv preprint arXiv:2505.09754}, 2025.

\bibitem[\protect\citeauthoryear{Kadavath and others}{2022}]{kadavath2022language}
Saurav Kadavath et~al.
\newblock Language models (mostly) know what they know.
\newblock {\em arXiv preprint arXiv:2207.05221}, 2022.

\bibitem[\protect\citeauthoryear{Kamath \bgroup \em et al.\egroup }{2023}]{kamath2023whatsup}
A.~Kamath, C.~Meng, J.~Li, L.~Zhou, J.~Hessel, Y.~Bisk, and J.~Lu.
\newblock What's "up" with vision-language models? investigating their struggle with spatial reasoning.
\newblock In {\em OpenReview}, 2023.

\bibitem[\protect\citeauthoryear{Khan and others}{2023}]{khan2023consistency}
Ofir~Press Khan et~al.
\newblock Measuring and narrowing the compositionality gap in language models.
\newblock {\em arXiv preprint arXiv:2210.03350}, 2023.

\bibitem[\protect\citeauthoryear{Li and others}{2023}]{li2023blip2}
Junnan Li et~al.
\newblock Blip-2: Bootstrapping language-image pre-training with frozen image encoders and large language models.
\newblock In {\em ICML}, 2023.

\bibitem[\protect\citeauthoryear{Liu and others}{2023}]{liu2023visual}
Fangyu Liu et~al.
\newblock Visual spatial reasoning.
\newblock In {\em TMLR}, 2023.

\bibitem[\protect\citeauthoryear{Liu \bgroup \em et al.\egroup }{2023}]{liu2023grounding}
Shilong Liu, Zhaoyang Zeng, Tianhe Ren, Feng Li, Hao Zhang, Jie Yang, Chunyuan Li, Jianwei Yang, Hang Su, Jun Zhu, et~al.
\newblock Grounding dino: Marrying dino with grounded pre-training for open-set object detection.
\newblock In {\em arXiv preprint arXiv:2303.05499}, 2023.

\bibitem[\protect\citeauthoryear{Manakul \bgroup \em et al.\egroup }{2023}]{manakul2023selfcheckgpt}
Potsawee Manakul, Adian Liusie, and Mark~JF Gales.
\newblock Selfcheckgpt: Zero-resource black-box hallucination detection for generative large language models.
\newblock {\em arXiv preprint arXiv:2303.08896}, 2023.

\bibitem[\protect\citeauthoryear{Radford and others}{2021}]{radford2021learning}
Alec Radford et~al.
\newblock Learning transferable visual models from natural language supervision.
\newblock In {\em ICML}, 2021.

\bibitem[\protect\citeauthoryear{Wang \bgroup \em et al.\egroup }{2025}]{wang2025spatialclip}
Z.~Wang, S.~Zhou, S.~He, H.~Huang, L.~Yang, Z.~Zhang, X.~Cheng, S.~Ji, T.~Jin, H.~Zhao, and Z.~Zhao.
\newblock Spatialclip: Learning 3d-aware image representations from spatially discriminative language.
\newblock In {\em Proceedings of the IEEE/CVF Conference on Computer Vision and Pattern Recognition (CVPR)}, pages 29656--29666, 2025.

\end{thebibliography}
}

\end{document}